%% file: main.tex
\pdfoutput=1

\documentclass[11pt]{article}

\usepackage[final]{acl}

\usepackage{times}
\usepackage{latexsym}

\usepackage[T1]{fontenc}

\usepackage[utf8]{inputenc}

\usepackage{microtype}

\usepackage{inconsolata}

\usepackage{graphicx}

\usepackage{xcolor} 
\definecolor{DarkGreen}{RGB}{1,50,32}

\usepackage{url}            

\usepackage{hyperref} 

\hypersetup{
    colorlinks=true,
    linkcolor=black,
    citecolor=blue,
    filecolor=cyan,
    urlcolor=blue
}

\usepackage{booktabs}
\usepackage{multirow}
\usepackage{colortbl}
\usepackage{caption}
\usepackage{subcaption}
\usepackage{algorithmic}
\usepackage[ruled,norelsize,linesnumbered]{algorithm2e}
\usepackage[most]{tcolorbox}
\usepackage[draft]{minted}
\usepackage{enumitem} 

\newcommand{\Autoref}[1]{%
  \begingroup%
  \def\algorithmautorefname{Algorithm}%
  \def\chapterautorefname{Chapter}%
  \def\sectionautorefname{Section}%
  \def\subsectionautorefname{Section}%
  \autoref{#1}%
  \endgroup%
}

\usepackage{scalerel,xparse}

\NewDocumentCommand\emojione{}{\raisebox{-0.1cm}{\scalerel*{\includegraphics{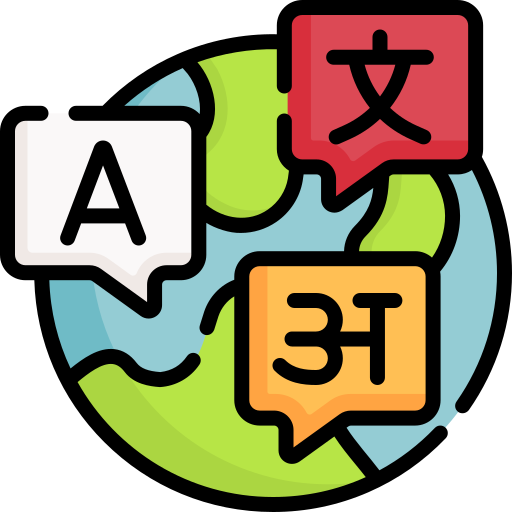}}{\rule{0.75cm}{0.75cm}}}}

\title{\vspace{-0.3cm}
\emojione\,Towards Fast Multilingual LLM Inference: \\Speculative Decoding and  Specialized Drafters}

\author{Euiin Yi${}^{1}$\thanks{Equal contribution.}\,\, Taehyeon Kim${}^{1*}$\,\, Hongseok Jeung${}^{2}$\,\, Du-Seong Chang${}^{2}$\,\, Se-Young Yun${}^{1}$  \\
  ${}^{1}$KAIST AI \quad  ${}^{2}$KT \\
  \texttt{\{euiin\_mercyii, potter32, yunseyoung\}@kaist.ac.kr} \\
  \texttt{\{hs.jeung,  dschang\}@kt.com} \\
  \url{https://github.com/Kthyeon/Multilingual-SpecBench}
}

\begin{document}
\maketitle
\input{section/0_abstract}

\input{section/1_intro}

\input{section/2_method}
\input{section/3_experiment}
\input{section/4_discussion}
\input{section/5_conclusion}

\section*{Limitations}\label{sec:limit}
Despite the improvement, our approach, requiring separate drafters for each language, introduces complexities in deployment, especially in multilingual settings. 
For instance, in environments where multiple languages are frequently interchanged, such as multinational corporations or global customer service platforms, the lack of an automated drafter selection system could hinder operational efficiency. 
Currently, our study focuses on independent drafters; however, examining systems that utilize interdependent models, similar to Eagle and Medusa, might offer insights into more interesting strategies. 
Additionally, while our findings are promising for translation tasks, expanding this methodology to other multilingual applications, like real-time multilingual generation or summarization, is essential to understand its broader applicability and uncover additional constraints.

This work primarily presents no direct ethical concerns. Further discussions are detailed in \autoref{sec:broad} and \autoref{sec:discuss}.

\section*{Acknowledgement}

This work was partly supported by Institute for Information \& communications Technology Promotion (IITP) grant funded by the Korea government (MSIT) [No.RS-2019-II190075, Artificial Intelligence Graduate School Program (KAIST), 10\%, No. RS-2024-00457882, AI Research Hub Project, 50\%, and No. 2022-0-00871, Development of AI Autonomy and Knowledge Enhancement for AI Agent Collaboration, 40\%].


\clearpage
\bibliography{ref}

\clearpage  
\appendix


\input{appendix/app1}

\end{document}

%% file: section/0_abstract.tex
\begin{abstract}
Large language models (LLMs) have revolutionized natural language processing and broadened their applicability across diverse commercial applications. However, the deployment of these models is constrained by high inference time in multilingual settings. To mitigate this challenge, this paper explores a training recipe of an assistant model in speculative decoding, which is leveraged to draft and-then its future tokens are verified by the target LLM. We show that language-specific draft models, optimized through a targeted \textit{pretrain-and-finetune} strategy, substantially brings a speedup in inference time compared to the previous methods. We validate these models across various languages in inference time, out-of-domain speedup, and GPT-4o evaluation.
\end{abstract}

%% file: section/1_intro.tex
\section{Introduction}
Large language models (LLMs) such as Gemini\,\cite{gemini}, GPT\,\cite{gpt4}, and Llama\,\cite{touvron2023llama} have remarkable success across various natural language processing tasks.
Their deployment in commercial settings has expanded to include applications such as coding assistance, writing support, conversational interfaces, and tools for search\,\cite{reid2024gemini}.
Despite their potential, the practical deployment of these models is often limited by prohibitively high inference time, particularly in multilingual contexts\,\cite{ahia2023all}. 
For example, character-level and byte-level models exhibit encoding length discrepancies exceeding fourfold for certain language pairs, resulting in significant disparities in cost and inference time available to different language communities\,\cite{petrov2024language}.
These challenges present substantial hurdles to scalable and cost-efficient applications of LLMs.

Speculative decoding, utilizing assistant models, has emerged as a promising strategy to improve LLM inference efficiency\,\cite{speculative_decode, speculative_decode2, spec_survey}, inspired by speculative execution\,\cite{burton1985speculative}.
This method drafts potential future tokens by leveraging a smaller model for the initial predictions. 
In parallel, these tokens are verified by the target LLM, ensuring only outputs aligned with the target LLM's predictions are accepted.
Recent efforts are focused on aligning these initial predictions with the target LLM’s outputs\,\cite{onspec, zhou2023distillspec}.
This involves advancing the training methods and modifying the architectural design of drafters\,\cite{specinfer, li2024eagle}.

\begin{figure}
  \centering
  \includegraphics[width=\linewidth]{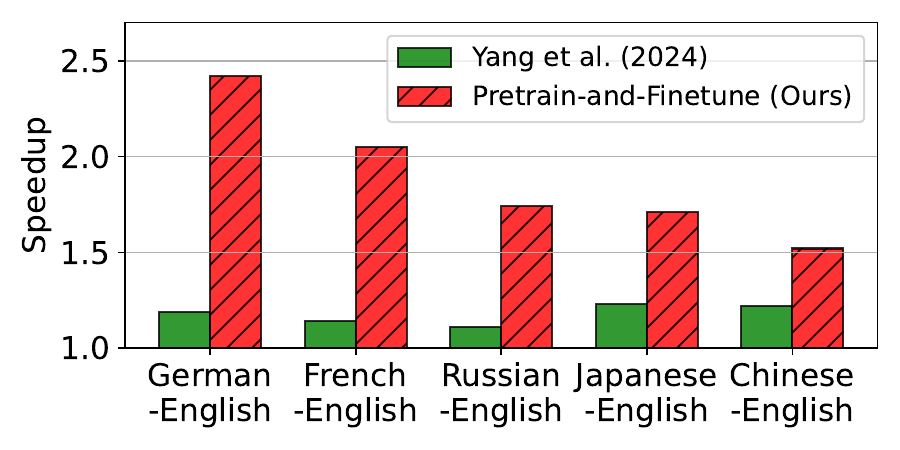}
  \caption{
  Speedup ratio\protect\footnotemark\, relative to the standard autoregressive greedy decoding on various multilingual datasets. Target model is Vicuna 7B v1.3 and the drafter is Vicuna 68M. Speculative greedy sampling is implemented with the drafter by \citet{vicuna68m} (\textcolor{DarkGreen}{green}) and our specialized drafter (\textit{pretrain-and-finetune}) (\textcolor{red}{red}). 
  }
  \label{fig:intro_motiv}
\end{figure}
\footnotetext{Evaluated on a single RTX3090 GPU with a batch size 1.}

Although speculative decoding has garnered considerable hype recently, the adaptation of this approach to multilingual scenarios common in real-world applications remains largely unexplored.
Prevailing methods\,\cite{medusa, li2024eagle, vicuna68m} use small drafters simply trained on datasets such as ShareGPT \cite{sharegpt} which is often used for instruction tuning of LLMs to learn a pattern of target LLM's language modeling.
However, our investigations reveal that such approaches are insufficient for multilingual translation (\autoref{fig:intro_motiv}).
This research also raises concerns regarding the capacity of such small drafters with simple tuning to comprehend the nuances of all target languages, thus questioning the feasibility of employing such models for universal speculative decoding.
This paper aims to shed light on the behaviors of drafters in speculative decoding within multilingual tasks and to explore their efficacy. 
Our contributions are as follows:

\begin{figure}
  \centering
  \includegraphics[width=\linewidth]{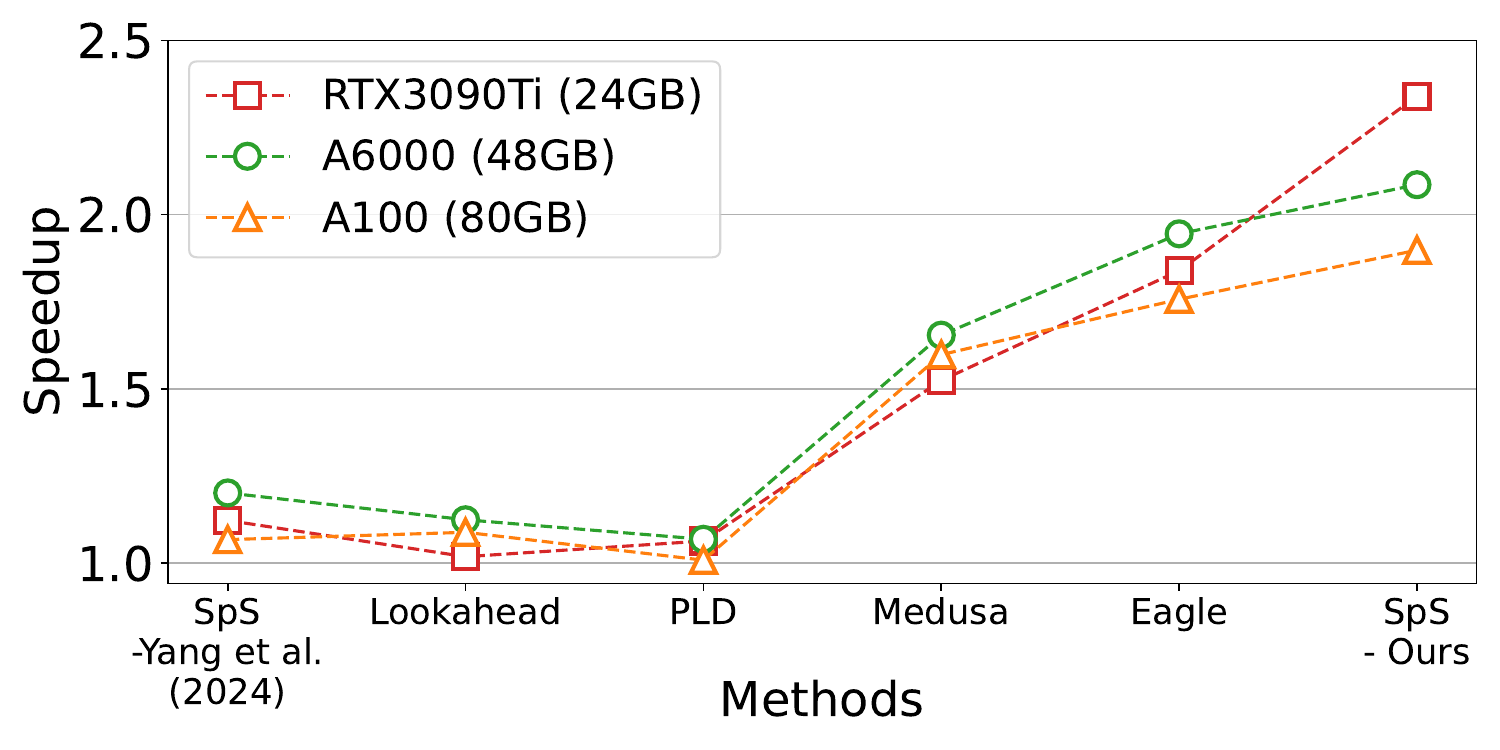}
  \caption{Speedup comparison of various speculative decoding methods on WMT16 De-En dataset\,\cite{wmt16} with greedy settings ($T$=$0.0$) across various hardwares. Target model is Vicuna-7B. 
  }
  \label{fig:intro}
\end{figure}

\begin{itemize}   
    \item We demonstrate that the strategy of \textit{pretrain-and-finetune} significantly improves the alignment of drafter models, achieving the highest speedup ratio among the baselines (\autoref{fig:intro}).

    \item We find that the speedup ratio increases as the number of tokens specific to the target task used in training increases. This speedup is logarithmically proportional to the scale of token count in drafter training.

    \item In multilingual translation, we observe that input languages consistent with the training set result in notable speedup, whereas outputs aligned with the training domain do not necessarily lead to improved performance. Additionally, our results are corroborated by GPT-4o judgment scores and qualitative analyses.

\end{itemize}

%% file: section/2_method.tex
\section{Method}
\subsection{Preliminaries: speculative decoding}
Speculative decoding employs a draft-verify-accept paradigm for fast inference. 
This method leverages a simpler assistant model ($M_p$) to predict easy tokens, thereby addressing memory bandwidth constraints in LLM inference\,\cite{trans_sampling}:

\begin{enumerate}
    \item \textbf{Draft}: An assistant model $M_p$, which is less computationally intensive than the target LLM $M_q$, drafts the future tokens $\{x_{t_1}, \ldots, x_{t_K}\}$ based on the input sequence $x_1, \ldots, x_t$.
    \item \textbf{Verify}: The target LLM $M_q$ assesses each token $x_{t_i}$ regarding whether it is aligned with its own predictions: $p_{i} = M_p(x_{t_i} | x_1, \ldots, x_t, x_{t_1}, \ldots, x_{t_{i-1}})$, $q_i = M_q(x_{t_i} | x_1, \ldots, x_t, x_{t_1}, \ldots, x_{t_{i-1}})$. 
    \item \textbf{Accept}:  Tokens meeting the validation criteria (e.g., rejection sampling) aligned with $M_q$'s outputs are retained. Tokens failing verification are either discarded or corrected, and the draft-verify cycle is repeated.
\end{enumerate} 

In this paper, the verification process employs rejection sampling\,\cite{speculative_decode, li2024eagle} when the temperature parameter is above zero to accept only tokens that closely match $M_q$'s predictions.
For greedy decoding with a temperature of zero, tokens are accepted if they are identical to $M_q$'s predictions.

\begin{figure}
  \centering
  \includegraphics[width=\linewidth]{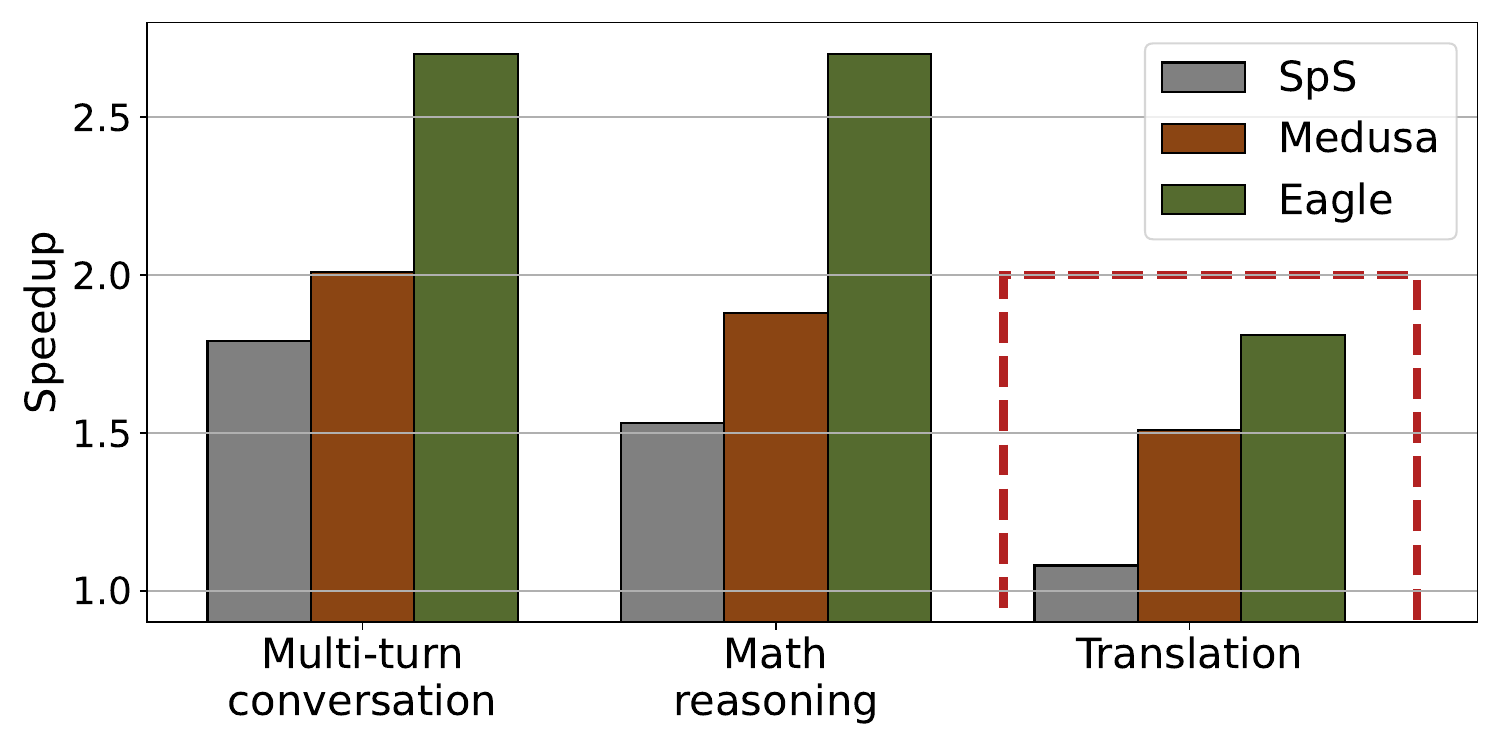}
  \caption{
  Speedup\protect\footnotemark\, comparison across categories containing multi-turn conversation (MT-Bench)\,\cite{llmjudge}, math reasoning (GSM8K)\,\cite{gsm8k}, and translation (WMT16 De-En). Target model is Vicuna-7B with speculative greedy sampling.
  }
  \label{fig:motivation}
\end{figure}
\footnotetext{Evaluated on a single RTX3090 GPU with a batch size 1.}

\subsection{Motivation}
Our evaluation of various speculative models, including SpS\,\cite{speculative_decode2}, Medusa\,\cite{medusa}, Eagle\,\cite{li2024eagle}, as shown in \autoref{fig:motivation}, demonstrates that speedup ratios significantly differ by task domain. 
While these models excel in English tasks such as multi-turn conversations and mathematical reasoning, where they achieve notable speed improvements, they underperform in translation tasks (red dotted box in \autoref{fig:motivation}). 
This result confirms that the effectiveness of these models is not universal but may be highly language-specific. 
The consistent underperformance in translation tasks highlights a key weakness and drives our study towards developing specialized drafters.

\subsection{Training specialized assistant models}

At the core of our approach is the recognition that smaller models, due to their inherent limited capacity, struggle to capture the diverse token distributions across languages. To address this challenge, we present specialized drafter models tailored to each language. Our strategy consists of:


\begin{enumerate}
    \item \textbf{Pretrain (P)}: Assistant models are pretrained on a part of C4\,\cite{c4} and ShareGPT dataset\,\cite{sharegpt} for language modeling.
    \item \textbf{Finetune (F)}: The models are finetuned on the target lingual task with instructions to refine their responses to non-English inputs.
\end{enumerate}

While the practices of pretraining and finetuning are well-established paradigms in language model training, applying these steps to drafter models represents a novel adaptation within the field. 
Traditionally, assistant models have been trained from scratch with little strategic rationale or clear justification for dataset selection. 

\begin{figure}
  \centering
  \includegraphics[width=\linewidth]{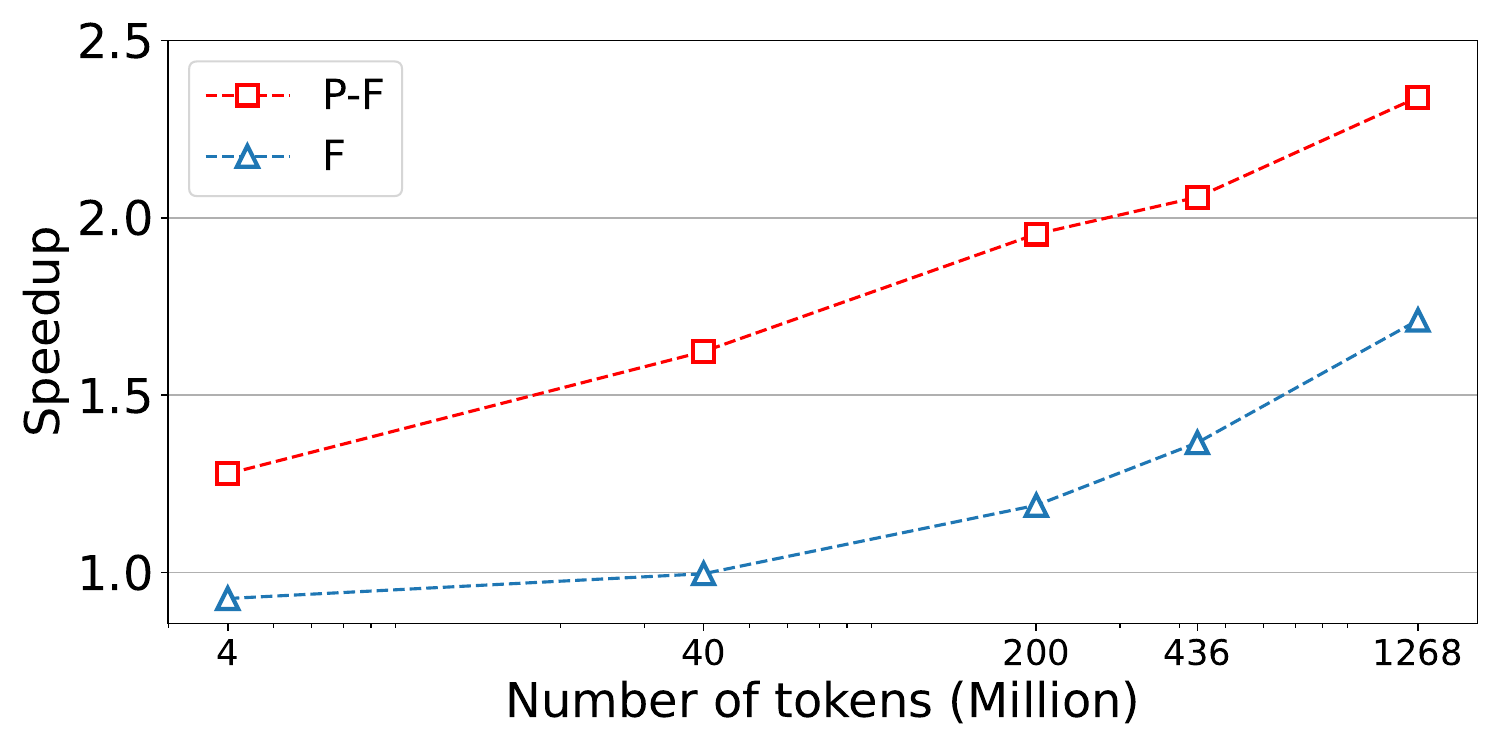}
  \caption{Speedup with speculative greedy sampling on the WMT16 De-En dataset as the training token for finetune (F) count varies, displayed on a logarithmic x-axis. `P-F' represents our strategy and `F' involves training solely on De-En without pretrain step (P).
  }
  \label{fig:strategy}
\end{figure}

\autoref{fig:strategy} shows that the \textit{pretrain-and-finetune} strategy significantly the speedup ratio as the number of training tokens increases. Our `P-F' approach outperforms models that are only finetuned (F), and even surpasses the speedup rates by \citet{vicuna68m}, which stood at 1.12.



\paragraph{Dataset with self-distillation} The training dataset for our assistant models is generated through self-distillation from the target LLM, ensuring alignment with its outputs\,\cite{seqKD, zhou2023distillspec, medusa}. 
To capture the full range of the target's output variability, we generate multiple responses at a range of temperatures—\{0.0, 0.3, 0.7, 1.0\}.

%% file: section/3_experiment.tex
\section{Experiment}\label{sec3}

\subsection{Experimental setup}

\paragraph{Models}
We utilize Vicuna 7B \cite{vicuna}, Gemma-Instruct 7B \cite{gemma}, and Llama2-chat\,\cite{touvron2023llama2} as target LLMs. The drafter models employed include Vicuna 68M \cite{vicuna68m}, a custom Gemma 250M drafter and Llama 68M\,\cite{specinfer}. Training configurations are outlined in \autoref{sec:imple}.

\paragraph{Number of drafts} 
For speculative sampling (SpS), we use a single draft candidate \cite{speculative_decode2}. In contrast, Medusa and Eagle models are evaluated using multiple drafts via tree-attention mechanism by following their original settings.

\paragraph{Training and evaluation}
Training datasets for each target model are generated via self-distillation and comprise five datasets: German (De)\textrightarrow English (En), French (Fr)\textrightarrow En, Russian (Ru)\textrightarrow En, Japanese (Ja)\textrightarrow En and Chinese (Zh)\textrightarrow En, each with 4 million (M) conversations ($\sim$ 1.3 billion (B) tokens) sourced from WMT14 Fr-En \cite{wmt14}, WMT16 De-En, and Ru-En \cite{wmt16}, and JParaCrawl-v3.0 \cite{japanese}. Evaluations are conducted using a single NVIDIA 3090 GPU, under both greedy settings ($T$=$0.0$) and for diversity at $T$=$1.0$ with three different seeds. 
The details are in \autoref{sec:imple}.

\begin{figure}
  \centering
  \includegraphics[width=\linewidth]{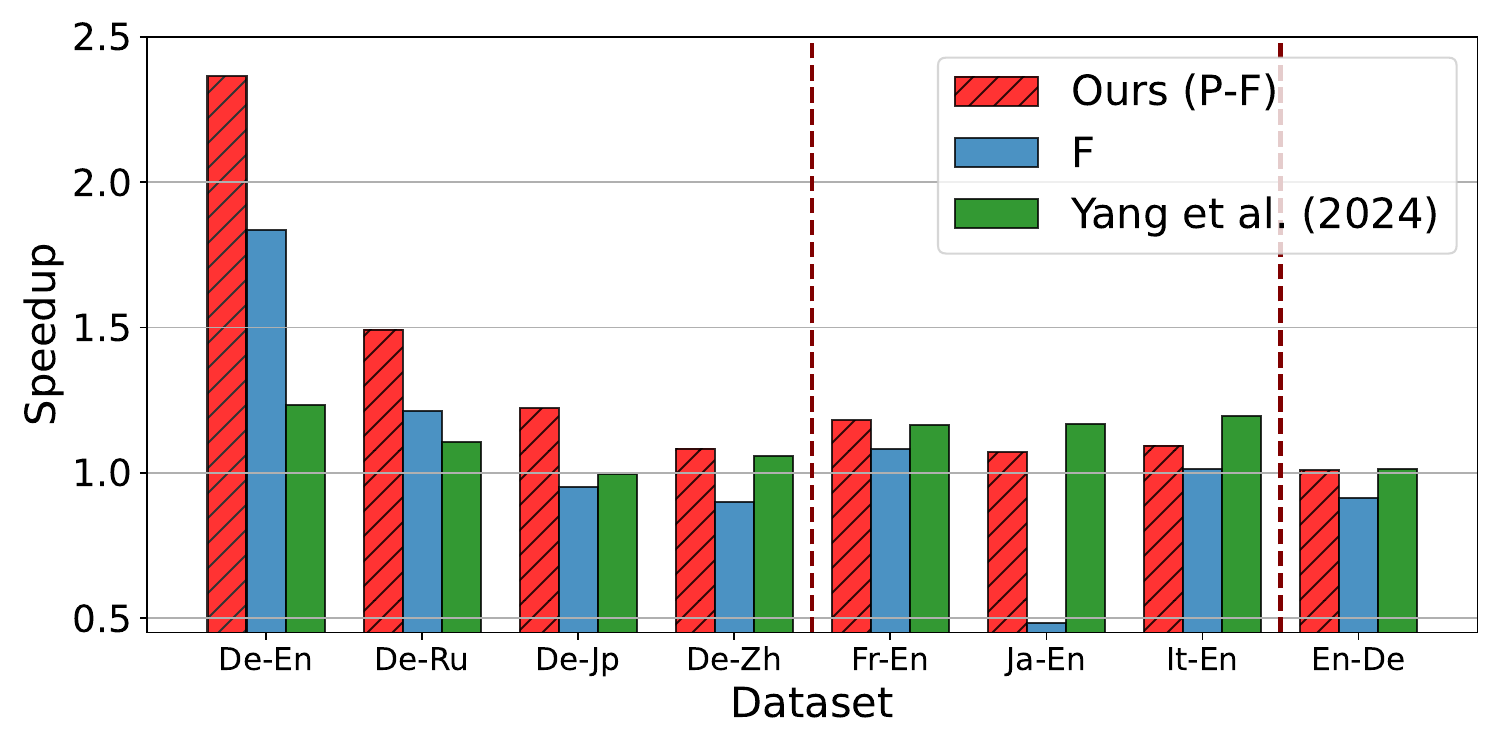}
  \caption{Speedup with speculative greedy sampling on various out-of-domain dataset as the drafters for `Ours (P-F)' and `F' are trained on WMT16 De-En dataset.}
  \label{fig:out-domain}
\end{figure}

\begingroup
\setlength{\tabcolsep}{6pt} 
\renewcommand{\arraystretch}{1.17}
\begin{table*}[t]
\caption{Speedup comparison of different methods for Vicuna 7B v1.3. Results are provided for $T$=$0.0$ and $T$=$1.0$ across various translation tasks. For our approach, each drafter is finetuned with the corresponding dataset.
\label{tab:main}
}
\resizebox{\textwidth}{!}{%
\begin{tabular}{@{}c|ccccc|c|ccccc|c@{}}
\toprule
\multirow{2}{*}{Method} & \multicolumn{6}{c|}{T=0.0}             & \multicolumn{6}{c}{T=1.0}             \\ \cmidrule(l){2-13} 
                        & De\textrightarrow En & Fr\textrightarrow En & Ru\textrightarrow En & Ja\textrightarrow En & Zh\textrightarrow En & Avg & De\textrightarrow En & Fr\textrightarrow En & Ru\textrightarrow En & Ja\textrightarrow En & Zh\textrightarrow En & Avg \\ \midrule\midrule
Sps - \citet{vicuna68m}           & 1.19$_{\pm 0.06}$  & 1.14$_{\pm 0.05}$  & 1.11$_{\pm 0.04}$  & 1.23$_{\pm 0.03}$ & 1.22$_{\pm 0.00}$ & 1.18$_{\pm 0.04}$ & 1.07$_{\pm 0.03}$  & 1.06$_{\pm 0.02}$  & 1.04$_{\pm 0.01}$      & 1.15$_{\pm 0.02}$ & 1.11$_{\pm 0.02}$ & 1.09$_{\pm 0.02}$    \\
Lookahead \cite{fu2024break}             & 1.03$_{\pm 0.01}$  & 1.01$_{\pm 0.02}$  & 0.98$_{\pm 0.01}$  & 1.00$_{\pm 0.01}$ & 0.96$_{\pm 0.00}$  & 1.00$_{\pm 0.01}$ & 1.03$_{\pm 0.03}$  & 1.04$_{\pm 0.03}$  & 0.99$_{\pm 0.00}$  & 0.98$_{\pm 0.05}$ & 0.98$_{\pm 0.00}$  & 1.01$_{\pm 0.02}$  \\
PLD \cite{saxena2023prompt}                   & 1.13$_{\pm 0.06}$  & 1.05$_{\pm 0.04}$  & 1.03$_{\pm 0.00}$  & 1.09$_{\pm 0.05}$ & 0.99$_{\pm 0.07}$ & 1.06$_{\pm 0.05}$  & -     & -     & -     & -     & -   & -   \\
Medusa \cite{medusa}                 & 1.58$_{\pm 0.05}$  &  1.57$_{\pm 0.01}$ & 1.52$_{\pm 0.01}$  & 1.55$_{\pm 0.01}$ & 1.43$_{\pm 0.00}$  & 1.53$_{\pm 0.02}$ & 1.61$_{\pm 0.03}$  & 1.69$_{\pm 0.01}$  & \textbf{1.62}$_{\mathbf{\pm 0.00}}$  & \textbf{1.72}$_{\mathbf{\pm 0.01}}$ & \textbf{1.60}$_{\mathbf{\pm 0.01}}$ & 1.65$_{\pm 0.01}$ \\
Eagle \cite{li2024eagle}                   & 1.90$_{\pm 0.05}$  & {1.88}$_{\pm 0.00}$  &  1.67$_{\pm 0.05}$  & \textbf{1.88}$\mathbf{_{\pm 0.01}}$ & \textbf{1.75}$_{\mathbf{\pm 0.01}}$ & 1.81$_{\pm 0.02}$ & 1.57$_{\pm 0.00}$  & 1.61$_{\pm 0.01}$      & {1.45}$_{\pm 0.02}$      & 1.63$_{\pm 0.01}$  & 1.51$_{\pm 0.03}$ & 1.55$_{\pm 0.01}$ \\
 \rowcolor{gray!20} Sps - \textit{pretrain-and-finetune} (Ours)                    & \textbf{2.42}$\mathbf{_{\pm 0.02}}$  & \textbf{2.05}$\mathbf{_{\pm 0.04}}$   & \textbf{1.74}$\mathbf{_{\pm 0.02}}$ & 1.71$_{\pm 0.01}$ & 1.52$_{\pm 0.01}$ &  \textbf{1.89}$\mathbf{_{\pm 0.02}}$ & \textbf{1.99}$\mathbf{_{\pm 0.01}}$ & \textbf{1.86}$\mathbf{_{\pm 0.03}}$      & {1.58}$\mathbf{_{\pm 0.00}}$   & 1.67$_{\pm 0.01}$  & 1.44$_{\pm 0.00}$ & \textbf{1.71}$_{\mathbf{\pm 0.01}}$ \\ \bottomrule
\end{tabular}
}
\end{table*}
\endgroup

\subsection{Main result}

\paragraph{Overall} 
\autoref{tab:main} shows that our specialized drafter (\textit{pretrain-and-finetune}) for targeted languages demonstrates superior performance across multiple translation tasks, recording the highest speedup in both deterministic ($T$=$0.0$) and diverse ($T$=$1.0$) settings. At $T$=$0.0$, our model outperforms all competitors with an average speedup ratio of 1.89. Similarly, at $T$=$1.0$, it maintains robust performance with an overall speedup ratio of 1.71.

\begin{table}[t] \scriptsize
\caption{Examples of speculative decoding on WMT16 De-En dataset. Black indicates standard decoded output and \textcolor{magenta}{magenta} indicates accepted draft tokens.}\label{tab:qualitative}
\centering
\begin{tabular}{@{}p{\linewidth}@{}}
\toprule
\textbf{Input} \\ \midrule
Translate German to English: So wie er gestartet ist , wird es nicht lange dauern , bis er auf der „ Pferd des Jahres “ -Schau ist – und ich bin mir sicher , dass er gut abschneiden wird. \\ \midrule
\textbf{SpS with a drafter by \citet{vicuna68m}} \\ \midrule
As he started, it \textcolor{magenta}{won'}t \textcolor{magenta}{take long} \textcolor{magenta}{until he}\textcolor{magenta}{'s} \textcolor{magenta}{on the} "\textcolor{magenta}{Horse} \textcolor{magenta}{of the} \textcolor{magenta}{Year"} show, and \textcolor{magenta}{I'}m \textcolor{magenta}{sure he'}ll do well.
\\\midrule
\textbf{Eagle} \cite{li2024eagle} \\ \midrule
\textcolor{magenta}{As he} \textcolor{magenta}{started,} \textcolor{magenta}{it won't} take \textcolor{magenta}{long until} \textcolor{magenta}{he'}\textcolor{magenta}{s on} the \textcolor{magenta}{"Horse of} \textcolor{magenta}{the Year"} \textcolor{magenta}{show,} \textcolor{magenta}{and I'm sure} \textcolor{magenta}{he'll do well}.
\\ \midrule
\textbf{SpS with our specialized drafter (\textit{pretrain-and-finetune})} \\ \midrule
\textcolor{magenta}{As he} \textcolor{magenta}{started, it won}\textcolor{magenta}{'t take long} \textcolor{magenta}{until he's} \textcolor{magenta}{on the "}Horse \textcolor{magenta}{of the} \textcolor{magenta}{Year"} show\textcolor{magenta}{, and} \textcolor{magenta}{I'm sure} \textcolor{magenta}{he'll} \textcolor{magenta}{do well.}
\\ \bottomrule
\end{tabular}
\end{table}

\paragraph{Speedup on out-of-domain translation tasks}
As \autoref{fig:out-domain} shows, our analysis reveals variability when applying the drafter, trained on the WMT16 De-En dataset, across diverse translation pairs.
Speedups are consistently higher when translating from German to other languages, highlighting the importance of input domain consistency with the training data. 
Conversely, translations involving non-German languages with English and English-German pairings show limited gains.
This result emphasizes that effective speculation depends more on matching the input domain of the translation task with the training data than on the output domain.

\paragraph{Qualitative analysis on responses}
\autoref{tab:qualitative} provides examples of speculative inference on the WMT16 De-En dataset. Both Eagle and our method incorporate a significant number of accepted tokens from drafts. However, our model achieves this with $\sim75\%$ fewer parameters, leading to reduced latency and faster inference time (\autoref{tab:main}). 
Similar to the findings in \citet{kim2024exploring}, Speculation typically takes place at critical junctions of the sentence such as transitions between clauses and phrases.

\begin{figure}[t]
    \centering
    \includegraphics[width=\linewidth]{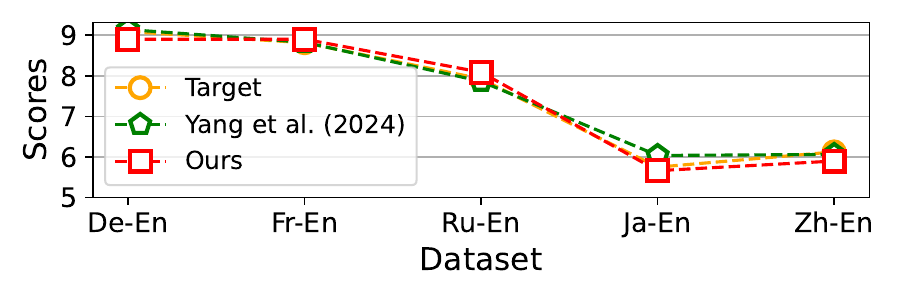}
    \caption{GPT-4o judgment scores following the \citet{llmjudge} on various multilingual translation dataset. The score is evaluated random sampling with $T$=$1.0$. }
    \label{fig:gpt4_judge}
\end{figure}

\begin{table}[t]
\centering
\caption{Ablations with speedup as the training stages continue on WMT19 Zh\textrightarrow En.}
\label{tab:ablation}\small
\resizebox{\linewidth}{!}{
\begin{tabular}{@{}l|cc@{}} 
\toprule
Target LLM - Drafter             & P & + F \\ \midrule
Gemma-Instruct 7B - Gemma 250M    & 0.92$_{\pm 0.01}$  & 1.04$_{\pm 0.02}$  \\
Llama2-chat 7B - Llama 68M & 1.47$_{\pm 0.00}$  & 1.95$_{\pm 0.01}$  \\ \bottomrule
\end{tabular}%
}
\end{table}

\paragraph{GPT-4o judgment analysis}
\autoref{fig:gpt4_judge} depicts the GPT-4o judgment scores\,\cite{llmjudge} generated using a temperature of 1.0. Our drafter closely matches the target Vicuna LLM across multiple datasets. The setup and further results are in \autoref{sec:imple} and \autoref{sec:additional}.


\paragraph{Ablation study} 
\autoref{tab:ablation} presents the ablation results, illustrating the progressive impact of the \textit{pretrain-and-finetune} approach on the performance of Gemma and Llama2-chat models.

%% file: section/4_discussion.tex
\section{Discussion}

\subsection{Why is pretrain-and-finetune better in small-size LM drafter?}
Drafting in speculative decoding has been treated akin to n-gram prediction  \cite{bhendawade2024speculative}, often relying on straightforward pretraining using datasets designed to replicate target LLM behaviors, such as the ShareGPT dataset \cite{vicuna68m}. This approach posits that generating a limited sequence of future tokens suffices for speculative inference.

Contrary to this belief, our empirical result presents a different narrative. \autoref{fig:out-domain} illustrates that even in seemingly straightforward translation tasks, such as from German to English, outcomes are not as effective. 
This suggests that drafting requires a broader array of language modeling capabilities to manage complex linguistic structures and context variations effectively.

Drafters, therefore, benefit significantly from a robust \textit{pretrain-and-finetune} approach, where they are first exposed to a wide array of linguistic contexts and then finely tuned to specific tasks. 
This training regimen transforms them into compact, yet comprehensive, language models capable of handling diverse and challenging speculative decoding scenarios with better alignment.

\subsection{Number of drafts}
This study primarily explores the speculative decoding process utilizing a single draft. In contrast, advanced baseline methods such as Eagle and Medusa deploy multiple drafts, leveraging tree-attention mechanisms to enrich draft selection. This technique allows for a broader exploration of multiple draft candidates at each decoding step, potentially increasing the rate and quality of accepted drafts.

Adapting our approach to incorporate multiple drafts with tree-attention could significantly enhance performance, suggesting an untapped potential in our method. Experimenting with this expanded setup could lead to notable improvements in the speculative sampling's effectiveness, particularly in increasing the mean number of high-quality tokens accepted per sequence. This prospect opens a critical path for future research, where deeper explorations could elevate the capabilities of our specialized drafters.

Further discussions are in \Autoref{sec:additional}.

%% file: section/5_conclusion.tex
\section{Conclusion}

This paper has demonstrated that the \textit{pretrain-and-finetune} strategy for training drafters significantly enhances speedup ratio relative to standard autoregressive decoding in multilingual translation tasks. 
This gain grows logarithmically with the increase in the number of training tokens.
Supported by qualitative analysis, out-of-domain analysis, and GPT-4o evaluation, this strategy substantially outperforms the state-of-the-art methods in various language pairs.
Our study uncovers approaches to maximize the benefits from drafter models, thereby setting a new benchmark in this area.

%% file: appendix/app1.tex
\section{Overview of appendix}\label{sec:appendix}
This appendix provides supplementary material that expands on the main contents. Each section is designed to complement the research presented:

\begin{itemize}
    \item \autoref{sec:broad}: \textbf{Broader impact} - Examines the wider implications of our findings on speculative decoding.

    \item \autoref{sec:future}: \textbf{Future work} - Outlines possible directions for future research, building upon the current study's findings to explore new avenues and improvements.

    \item \autoref{sec:related}: \textbf{Related works} - Provides a comprehensive review of literature and previous research that relate to the speculative decoding techniques discussed in the paper.

    \item \autoref{sec:algo}: \textbf{Algorithm} - Details the algorithms used in the speculative decoding processes, providing pseudocode and explanations to support reproducibility.

    \item \autoref{sec:imple}: \textbf{Implementation details} - Offers an in-depth look at the practical implementation of the speculative decoding methods, including baselines, self-distillation, training, and GPT-4o evaluation.

    \item \autoref{sec:additional}: \textbf{Additional experimental results} - Presents extra experimental data and analyses that were not included in the main sections due to space constraints.

    \item \autoref{sec:discuss}: \textbf{Discussions} - Engages in discussions on results, such as foundational beliefs that underpin our research approach, the number of drafts used, and drafter size.
\end{itemize}

Each appendix is intended to provide clarity and additional context to the research.

\section{Broader impact}\label{sec:broad}
Implementing language-specific drafters significantly enhances the speed of large language models tailored to diverse linguistic environments. 
For instance, a system could leverage heuristic analysis of input prompt token distributions to automatically select an optimal drafter, streamlining processing efficiency. 
Moreover, if a user interface allows individuals to choose their preferred language, the system can instantly apply the corresponding drafter, thereby accelerating response times considerably. 
Such advancements not only reduce computational load but also enrich the user experience by providing rapid and culturally relevant responses in multilingual contexts.

\section{Future work}\label{sec:future}
Future projects will explore broadening the scope of our speculative decoding framework to cover general multi-task environments, extending beyond multilingual translation to include varied domains such as legal and medical text processing. 
A significant challenge lies in developing an efficient method for selecting the appropriate drafter among multiple options when direct user input is unavailable or when inputs consist of mixed languages. 
This issue becomes more complex as the ambiguity of language indicators increases. 
To alleviate this, designing an advanced router capable of intelligently assigning tasks to the most suitable drafter based on the nature of the input presents a promising direction. 
Training such a router involves leveraging advanced techniques to understand and predict the optimal drafter based on contextual representations. 
This approach aims to improve the overall efficiency and accuracy of language model applications across diverse and dynamically changing content landscapes.

\begin{algorithm*}
\DontPrintSemicolon
    \caption{Speculative sampling}
    \label{alg:sps}
\begin{algorithmic}[1]
    \SetKwInOut{Input}{Input}
    \INPUT: Target LLM $\mathcal{M}_q$, a small assistant model $\mathcal{M}_p$, initial prompt sequence $x_1, \ldots, x_t$ and target sequence length $T$.\\
    \STATE Initialize $t \leftarrow 1$ \\
    \WHILE{$t < T$} 
        \FOR{$k \leftarrow 1, \ldots, K$}
            \STATE $x_{t_k} \sim \mathcal{M}_p(x|x_1, \ldots, x_t, x_{t_1}, \ldots, x_{t_{k-1}})$
        \ENDFOR
        \STATE In parallel, compute $K+1$ sets of logits drafts $x_{t_1}, \ldots, x_{t_K}$ with the target LLM $\mathcal{M}_q$: \\
         \quad \quad $\mathcal{M}_q(x|x_1, \ldots, x_t), \mathcal{M}_q(x|x_1, \ldots, x_t, x_{t_1}), \ldots, \mathcal{M}_q(x|x_1, \ldots, x_t, x_{t_1}, \ldots, x_{t_K})$
        \FOR{$j \leftarrow 1, \ldots, K$}
            \STATE Sample $r \sim U[0,1]$ from a uniform distribution
            \IF{$r < \min(1, \frac{\mathcal{M}_q(x|x_1, \ldots, x_{t+j-1})}{\mathcal{M}_p(x|
            x_1, \ldots, x_{t+j-1})})$}
                \STATE Set $x_{t+j} \leftarrow x_{t_{j}}$ and $t \leftarrow t +1$
            \ELSE
                \STATE Sample $x_{t+j} \sim (\mathcal{M}_q(x|x_1, \ldots, x_{t+j-1}) - \mathcal{M}_p(x|x_1, \ldots, x_{t+j-1}))_{+}$ and exit for loop.
            \ENDIF
        \ENDFOR
        \STATE If all tokens $x_{t+1}, \ldots, x_{t+K}$ are accepted, sample extra token\\  $x_{t+K+1} \sim \mathcal{M}_q(x|x_1, \ldots, x_t, x_{t+K})$ and set $t \leftarrow t+1$
    \ENDWHILE
\end{algorithmic}
\end{algorithm*}

\section{Related works}\label{sec:related}

\subsection{Speculative decoding}

Speculative decoding, advancing from blockwise parallel decoding introduced by \citet{bpd}, adopts a draft-then-verify paradigm to enhance LLM inference efficiency. 
This method addresses latency issues in autoregressive decoding, which stem from the extensive memory transfers required for each token generation, leading to computational underutilization \cite{spec_survey, patterson2004latency}.
To further advance this paradigm, \citet{speculative_decode} and \citet{speculative_decode2} introduced speculative decoding and sampling, which includes the lossless acceleration of various sampling methods. These methods utilize smaller models from the same series, such as T5-small, to accelerate inference for larger counterparts like T5-XXL without additional training.

Recent advancements in speculative decoding, exemplified by models like EAGLE \cite{li2024eagle} and Medusa \cite{medusa}, have significantly refined the efficiency of LLMs by integrating lightweight feedforward neural network (FFN) heads directly into their architecture. 
These FFN heads facilitate the early drafting of token sequences, enhancing throughput and reducing latency. 
Similarly, approaches such as the self-speculative model \cite{zhang2023draft} and \citet{elhoushi2024layer} incorporate early exiting and layer skipping strategies, allowing for a reduction in computational load by prematurely terminating decoding processes or bypassing less impactful neural layers. 
Another line of research explores the blockwise parallel language models with multiple softmax heads pretrained from scratch presented by \citet{bpd} by either refining its drafts \cite{kim2024exploring} or scaling up the model size \cite{gloeckle2024better}.

\subsection{Inference acceleration of LLM}
As LLMs continue to evolve rapidly, enhancing their inference speed has become a focal area of research. Traditional techniques such as knowledge distillation \cite{gu2023minillm, ko2024distillm}, model compression \cite{li2020train}, and quantization \cite{xiao2023smoothquant} aim to optimize these models but often require extensive training adjustments or significant architectural modifications. More recent strategies have shifted towards applying early exiting mechanisms, particularly within series like T5 \cite{calm, bae2023fast} and decoder-only architectures \cite{varshney2023accelerating}, to streamline inference processes. Although early exiting can significantly hasten model responses by truncating computational sequences, this method typically introduces a trade-off with performance degradation \cite{calm}.

\section{Algorithm: speculative sampling} \label{sec:algo}
By referring to \citet{speculative_decode2}, \Autoref{alg:sps} demonstrates the speculative sampling process. Initiating with an initial prompt, an assistant model is utilized to generate multiple prospective continuations at each step, which are concurrently verified against the target LLM’s predictions.

Each candidate token's acceptance probability is calculated based on the target LLM's relative confidence compared to the assistant model's suggestion (i.e., rejection sampling). If a value, randomly drawn from a uniform distribution, falls below this threshold, the token is accepted and incorporated into the ongoing sequence. If not, the algorithm recalibrates, adjusting the speculative path by directly sampling from the differences in predictions, enhancing accuracy and contextual relevance.

\section{Implementation details}\label{sec:imple}

\subsection{Baselines}
Following the Spec-Bench settings \cite{spec_survey}, we have selected 5 speculative decoding methods, all open-source and rigorously tested for reliability. Each method represents a unique approach to improving LLM inference speeds:

\begin{enumerate}
    \item \textbf{SpS} \cite{speculative_decode2}: SpS employs a smaller LM from the same model series as the drafter. In the verification, this method corrects the last token with residual probability if the token is rejected.
    \item \textbf{Medusa} \cite{medusa} and \textbf{Eagle} \cite{li2024eagle}: Both methods enhance the target LLM by integrating additional lightweight FFN heads. These heads are designed to efficiently draft potential token sequences depending on the penultimate representations from the target LLM.
    \item \textbf{Lookahead} \cite{fu2024break}: This method appends multiple special tokens to the end of the input prompt. These tokens are used for parallel drafting, with the resultant drafts transformed into n-gram candidates for efficient prediction.
    \item \textbf{PLD} \cite{saxena2023prompt}: Serving as the practical code implementation of \citet{yang2023inference}, PLD selects text spans directly from the input to serve as drafts, optimizing the relevance and accuracy of the initial predictions.
\end{enumerate}

\subsection{Self-distillation}
We follow the self-distillation pipeline as described by \citet{medusa}. 
Initially, a public dataset, such as WMT 16 De-En, is selected as the training dataset. The target model's responses are then generated using the OpenAI API server, with input prompts derived directly from the training dataset.

\paragraph{Install prerequisites} 
For software dependencies, CUDA 12.1 and PyTorch 2.1.2 are required. To start the server, install the necessary dependencies:

\begin{minted}{objc}
      vllm==0.4.0, openai==0.28.0
\end{minted}  

\paragraph{Use of vLLM} 
We utilize the vLLM library for self-distillation, executing the following command:
\definecolor{LightGray}{gray}{0.7}
\begin{minted}[frame = single]{latex}
python -m 
vllm.entrypoints.openai.api_server
--model lmsys/vicuna-7b-v1.3
--port 8000 --max-model-len 2048
\end{minted}

\paragraph{Input prompt} 
For instance, when self-distillation the WMT14 Fr-En dataset using the Vicuna7b v1.3 model, the input prompt consists of a system prompt and a user prompt. In the user prompt, we prepend "Translate French to English: ".
\begin{tcolorbox} [colback=gray!5!white,colframe=gray!75!black]
A chat between a curious user and an artificial intelligence assistant. The assistant gives helpful, detailed, and polite answers to the user's questions. USER: Translate French to English: Madame la Présidente, c'est une motion de procédure. ASSISTANT:
\end{tcolorbox}

\subsection{Details on training setup}
For the shared settings across all training drafters, we employ the Fastchat\footnote{\url{https://github.com/lm-sys/FastChat/tree/main}} framework. We utilize a cosine learning rate scheduler with a warmup ratio of 0.03 and the AdamW\,\cite{adamw} optimizer. 
The drafter is trained using the `P-F' strategy (ours) for 3 epochs, and using the `F' strategy (without the pretraining step `P') for 5 epochs to ensure sufficient learning. The model's maximum length is set to 2048 tokens. The training is conducted using 4 GPUs with a batch size of 2 per GPU.

For finetuning the Vicuna 68M drafter\,\cite{vicuna68m}, the learning rate is set to 2e-5. Similarly, for finetuning the Llama 68M model\,\cite{specinfer}, the learning rate is set to 3e-5. 

As a drafter for Gemma-Instruct 7B model, we newly design a Gemma 250M model as a drafter (\autoref{tab:gemma}). We use the same training recipe with Vicuna 68M and Llama 68M.

\begin{table}[t]\small
\caption{Custom Gemma 250M model configuration.}
\label{tab:gemma}
\centering
\resizebox{\linewidth}{!}{%
\begin{tabular}{@{}c|c@{}}
\toprule
Configuration & Value \\ \midrule
Activation function       & GeLU \cite{hendrycks2016gaussian}   \\ 
Hidden size               & 768    \\
Intermediate size         & 6144   \\
Number of attention heads & 16     \\
Number of hidden layers   & 2      \\
Number of key-value heads & 2      \\
RMS epsilon               & 1e-06  \\
Vocabulary size           & 256000 \\ \bottomrule
\end{tabular}%
}
\end{table}

\subsection{Details on GPT-4o evaluation}
We follow LLM-as-a-Judge framework \cite{llmjudge} to evaluate the model's answers. The GPT-4o model is utilized as a judge, which has greater performance on both English and non-English than GPT-4 Turbo \cite{openai2024gpt4o}. For \textbf{Single answer grading}, used prompt is followed:

\begin{tcolorbox} [colback=gray!5!white,colframe=gray!75!black]
[System] \\
You are a helpful assistant. Please act as an impartial judge and evaluate the quality of the response provided by an AI assistant to the user question displayed below. Your evaluation should consider factors such as the helpfulness, relevance, accuracy, depth, creativity, and level of detail of the response. Begin your evaluation by providing a short explanation. Be as objective as possible. After providing your explanation, you must rate the response on a scale of 1 to 10 by strictly following this format: "[[rating]]", for example:  "Rating: [[5]]".
\\
\\
{[Question]} \\
\{question\}
\\
\\
{[The Start of Assistant's Answer]} \\
\{answer\} 

[The End of Assistant's Answer]
\end{tcolorbox}

The detail implementation of LLM-as-a-judge is in the following GitHub repository\footnote{\url{https://github.com/lm-sys/FastChat/tree/main/fastchat/llm_judge}}.

\begin{figure*}[!ht]
    \centering
    \begin{subfigure}{0.49\textwidth}
        \includegraphics[width=\linewidth]{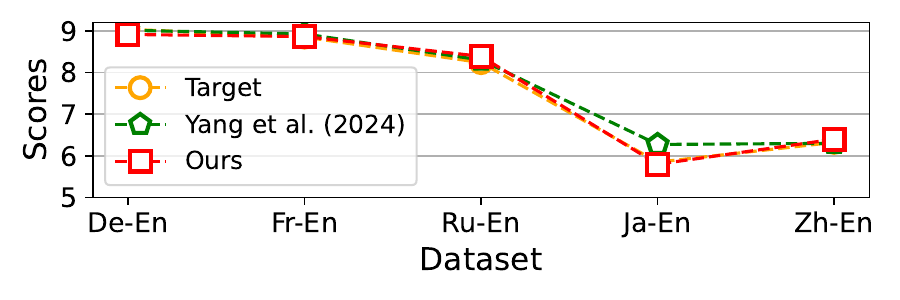}
        \caption{T=0.8}    
    \end{subfigure}
    \begin{subfigure}{0.49\textwidth}
        \includegraphics[width=\linewidth]{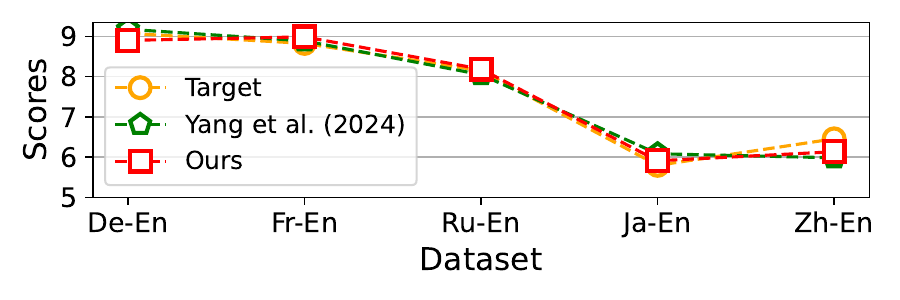}
        \caption{T=0.9}    
    \end{subfigure}
    \caption{GPT-4o evaluation scores following the \citet{llmjudge} on various multilingual translation dataset. Each figure denotes the score of random sampling with different temperature on the output whose target LLM is Vicuna 7B v1.3.}
    \label{fig:add_temp}
\end{figure*}

\begin{figure*}[!ht]
    \centering
    \begin{subfigure}{0.325\textwidth}
        \includegraphics[width=\linewidth]{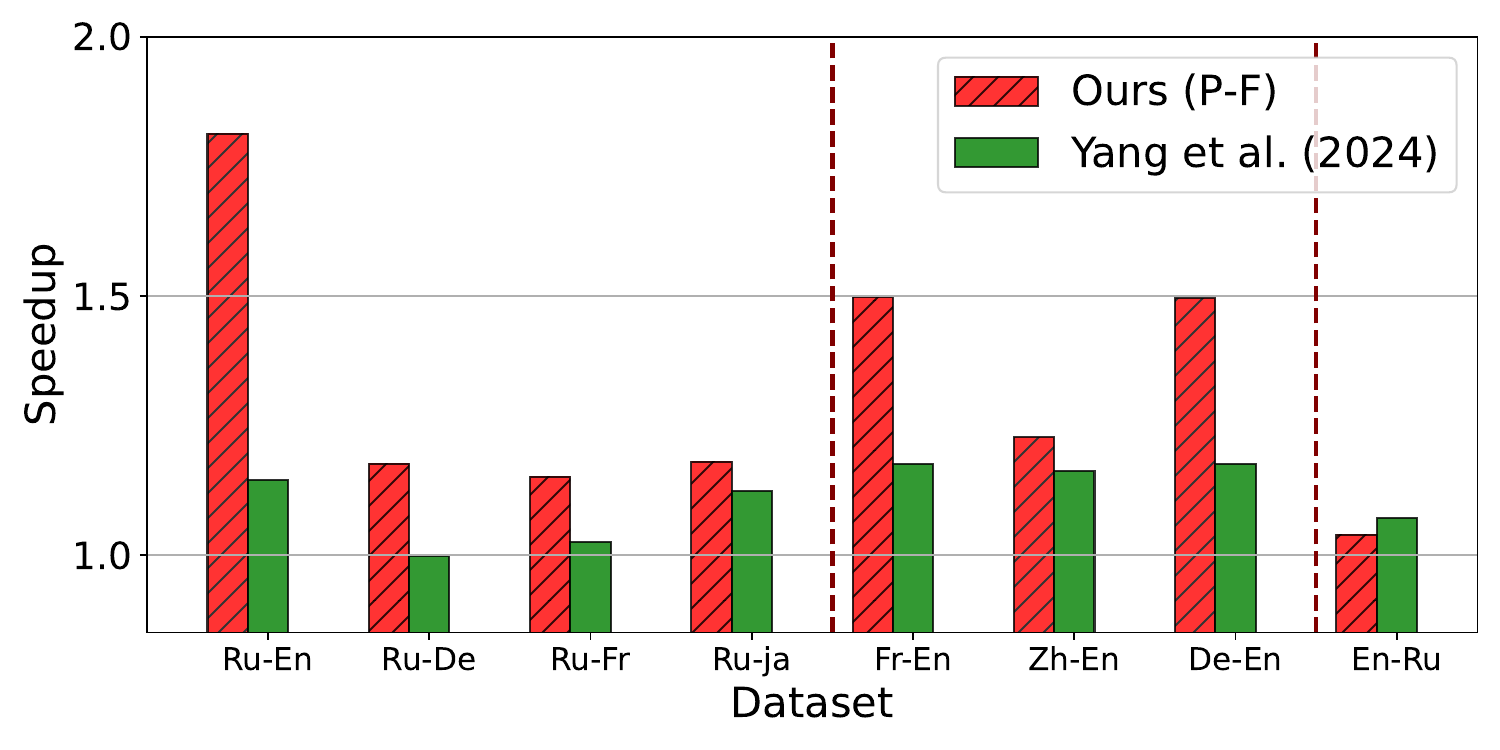}
        \caption{Drafter trained on Ru-En}
        \label{fig:outdomain_Ru-En}
    \end{subfigure}
    \begin{subfigure}{0.325\textwidth}
        \includegraphics[width=\linewidth]{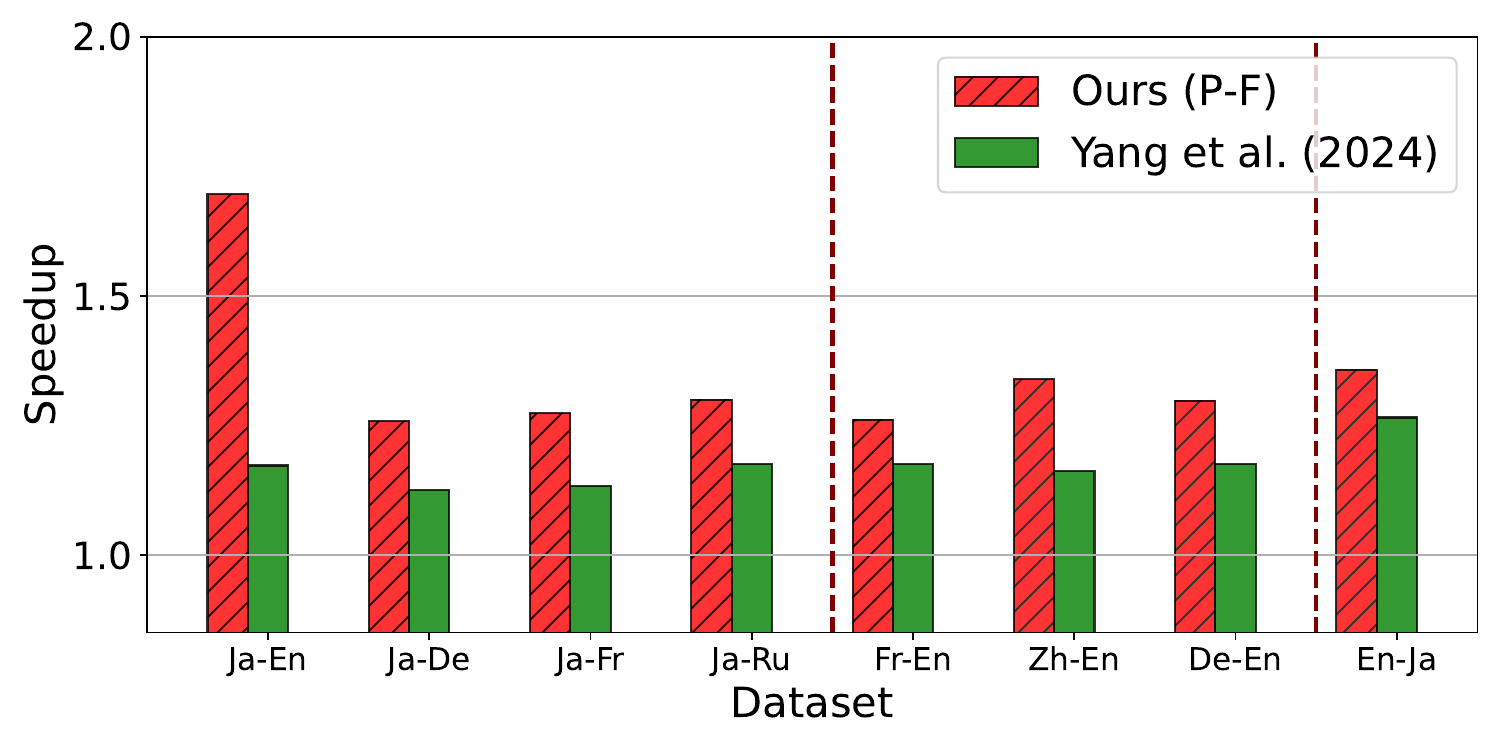}
        \caption{Drafter trained on Ja-En}    \label{fig:outdomain_Ja-En}   
    \end{subfigure}
    \begin{subfigure}{0.325\textwidth}
        \includegraphics[width=\linewidth]{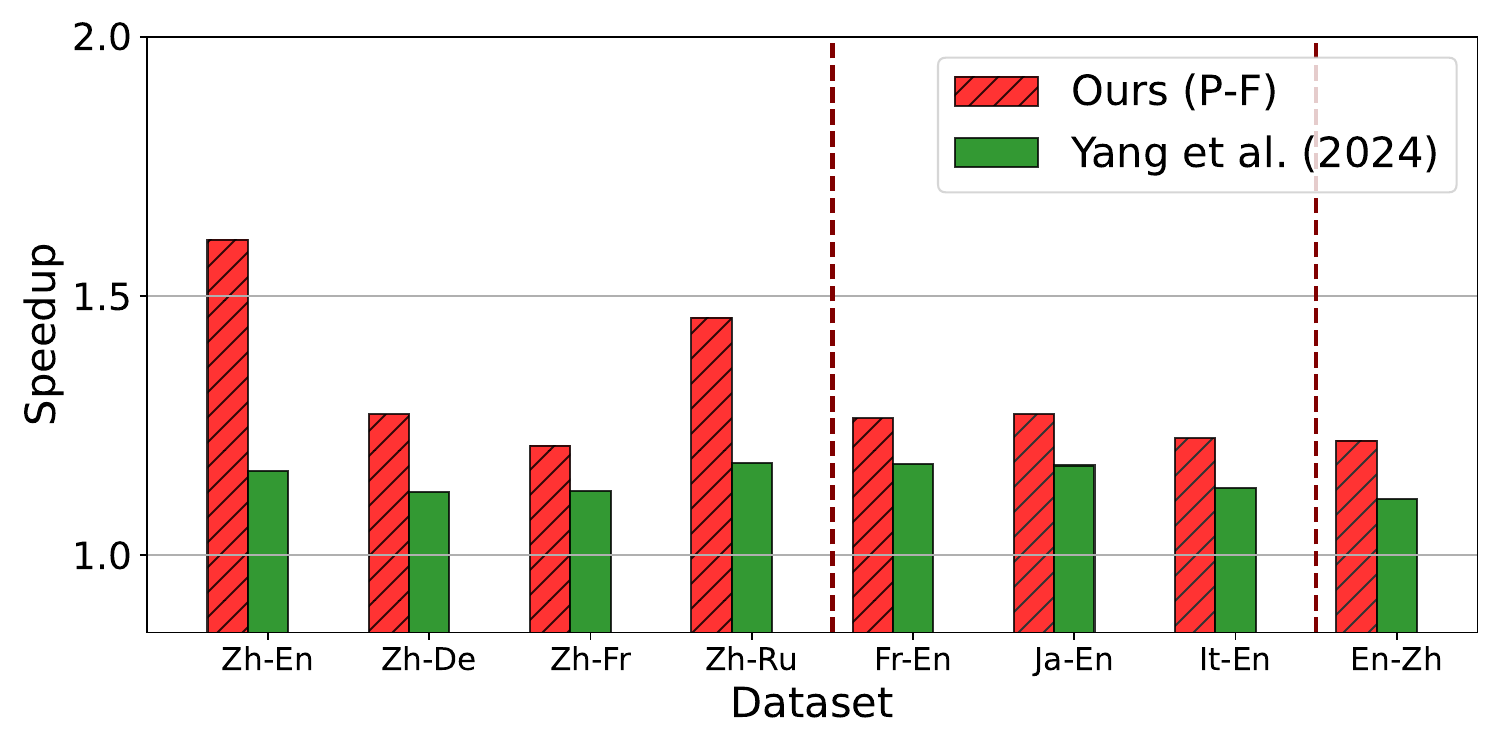}
        \caption{Drafter trained on Zh-En}    \label{fig:outdomain_Zh-En}
    \end{subfigure}
    \caption{Speedup with speculative greedy sampling with the same settings in \autoref{fig:out-domain}.}
    \label{fig:add_outdomain}
\end{figure*}

\section{Additional experimental results}\label{sec:additional}

\subsection{Average acceptance length comparison}
Building on the main findings, we further explore average acceptance length, a hardware-agnostic metric that measures the number of tokens accepted from a draft or generated per drafting-verification cycle. The key advantage of average acceptance length is its independence from hardware and runtime environments. However, its limitation lies in its inability to account for the overhead introduced by the draft model.\autoref{tab:main-average_acceptance_length} shows average acceptance length for different methods on De-En translation tasks across $T=0.0$ and $T=1.0$.

Our method, Sps with pretrain-and-finetune, achieved 3.03 at $T=0.0$ and 2.50 at $T=1.0$, outperforming traditional methods like Sps (\cite{vicuna68m}) and Lookahead, which reached 1.47 and 1.23, respectively. Even compared to self-drafting methods like Medusa and Eagle, our approach remained competitive, demonstrating the effectiveness of our strategy in improving block acceptance rates.

These results highlight the efficiency of our method in accepting more tokens per draft, leading to faster, more efficient processing across diverse datasets.

\begin{table}[t]\small
\caption{Average acceptance length comparison of different methods for Vicuna 7B v1.3. Results are provided for $T$=$0.0$ and $T$=$1.0$ across De\textrightarrow En translation tasks.}
\label{tab:main-average_acceptance_length}
\centering
\resizebox{\linewidth}{!}{%
\begin{tabular}{@{}c|c|c@{}}
\toprule
Method & T=0.0 & T=1.0 \\ \midrule
Sps - \citet{vicuna68m}           & 1.47  & 1.35   \\
Lookahead \cite{fu2024break}             & 1.23  & 1.23   \\
PLD \cite{saxena2023prompt}                   & 1.15  & -   \\
Medusa \cite{medusa}                 & 2.22  & 2.29   \\
Eagle \cite{li2024eagle}                   & 3.04  & 2.70   \\
 \rowcolor{gray!20} Sps - \textit{pretrain-and-finetune} (Ours)  & 3.03  & 2.50 \\ \bottomrule
\end{tabular}%
}
\end{table}

\subsection{Out-of-domain speedup}
Building on the findings discussed in the main body, this subsection further explores the speedup variations achieved by employing a drafter trained on each dataset across a range of translation tasks. \autoref{fig:add_outdomain} depicts the speedup results using speculative greedy sampling for drafters trained on different datasets: Ru-En, Ja-En, and Zh-En. 

Most observations align with those discussed in \Autoref{sec3}. Notably, drafters trained on the Ja-En (\autoref{fig:add_outdomain}\,(\subref{fig:outdomain_Ja-En})) and Zh-En (\autoref{fig:add_outdomain}\,(\subref{fig:outdomain_Zh-En})) datasets consistently outperform \citet{vicuna68m}'s drafter, even on out-of-domain tasks. 
We hypothesize these into two folds. Firstly, this suggests that certain intrinsic properties of the Japanese and Chinese languages may improve the efficacy of speculative decoding when applied to unrelated language pairs, possibly due to specific syntactic or lexical features that are effectively captured during training. 
In another scenario, the target LLM does not work well on those tasks, and thus drafters are easier to catch the target token distribution. 
More precisely, for instance, in Zh-Ru task, Vicuna 7B should translate the Chinese to Russian, but to English, and thus the speedup seems to happen for us due to English generation. 

In the case of the Ru-En (\autoref{fig:add_outdomain}\,(\subref{fig:outdomain_Ru-En})) trained drafter, translations from Russian to other languages generally surpass \citet{vicuna68m}'s results.
Interestingly, translations from French to English and German to English exhibit unexpectedly high speedups. 
This could hint at underlying linguistic similarities or shared grammatical structures between Russian, French, and German that the Ru-En drafter is particularly adept at handling, thereby facilitating more efficient speculative decoding. While \citet{fan2021discovering} demonstrates that Russian belongs to another cluster from En / Fr / De, perhaps our results provide a different perspective in lens of speculative decoding.

\begin{table*}[t]
\caption{Speedup comparison of speculative greedy sampling across different drafter sizes on WMT16 De-En dataset.}
\label{tab:drafter_discuss}
\resizebox{\linewidth}{!}{%
\begin{tabular}{@{}l|ccc@{}}
\toprule
Drafter                 & Vicuna 68M \cite{vicuna68m} & Vicuna 68M (\textit{pretrain-and-finetune}; Ours) & Tiny-Vicuna 1B \cite{tinyvicuna1b} \\ \midrule
Speedup                 & 1.19                     & \textbf{2.42}              & 0.75      \\
Mean of accepted tokens & 1.47                     & 3.03              & \textbf{3.06}      \\ \bottomrule
\end{tabular}%
}
\end{table*}

\begin{table*}[t]
\caption{Speedup results for same language pairs, different datasets.}
\label{tab:ood_data}
\resizebox{\linewidth}{!}{%
\begin{tabular}{@{}l|cc@{}}
\toprule
Model           & Speedup (WMT16 De-En fine-tune, WMT16 De-En eval) & Speedup (WMT16 De-En fine-tune, IWSLT14 De-En eval) \\ \midrule
Sps-\cite{vicuna68m}          & 1.19                                                        & 1.23                                                         \\
Sps - pretrain-and-finetune (Ours) & 2.42                                       & 2.51                                                         \\ \bottomrule
\end{tabular}%
}
\end{table*}

\subsection{GPT-4o judgments} 
\autoref{fig:add_temp} show additional GPT-4o evaluation scores for various multilingual translation datasets. The graphs display the comparative performance across different language pairs under two sampling conditions, at temperatures $T$=$0.8$ and $T$=$0.9$, respectively. Each data point reflects the quality of translations produced by the target model (\textcolor{orange}{orange circle}), SpS with the instruction tuned model using ShareGPT \cite{vicuna68m} (\textcolor{DarkGreen}{green pentagon}), and SpS with our specialized drafter (\textit{pretrain-and-finetune}) (\textcolor{red}{red square}).
For the red points, each drafter is trained with the corresponding dataset. For instance, when the red point specify De-En, it indicates that the drafter has been fine-tuned with the De-En dataset.

The results demonstrate negligible differences in quality among the three methods, underscoring the efficacy of speculative decoding in delivering translations with lossless quality. Both temperature settings show that our speculative decoding strategy closely matches the performance of the established target model across various language pairs. This consistent performance across different settings and language pairs illustrates that speculative decoding effectively maintains high-quality outputs without compromising accuracy due to increased randomness in sampling.

\section{Discussion} \label{sec:discuss}

\subsection{Is scaling up drafter size better for SpS?}

Evaluating the efficacy of increasing drafter size reveals nuanced insights into speculative decoding performance. \autoref{tab:drafter_discuss} compares three versions of drafters: the Vicuna 68M by \citet{vicuna68m}, our \textit{pretrain-and-finetune} Vicuna 68M, and Tiny-Vicuna 1B\,\cite{tinyvicuna1b}—a larger model with 1B parameters that has been instruction-tuned.

Despite Tiny-Vicuna 1B's substantial parameter count, it achieves a lower speedup of 0.75 compared to 2.34 by our optimized Vicuna 68M. Both models show similar mean accepted tokens, suggesting that increasing size does not proportionally enhance computational efficiency. This is due to speculative decoding's reliance on minimizing memory bottlenecks to exploit parallel computation effectively. Larger models like Tiny-Vicuna 1B exacerbate these bottlenecks, diminishing the potential speed gains from increased parallelism.

Conversely, our \textit{pretrain-and-finetune} Vicuna 68M demonstrates that strategic training and optimization of a smaller model can achieve high efficiency and speed, highlighting the importance of model configuration over mere size increase. This balance between model size and computational dynamics is crucial for optimizing speculative decoding, suggesting that enhancing model capabilities through targeted training may be more effective than scaling size.

\subsection{Evaluating generalization across datasets}
We fine-tune the model on WMT16 De-En and evaluated it on IWSLT14 De-En. As presented in \Autoref{tab:ood_data}, our specialized drafter demonstrate a speedup ratio of 2.51, surpassing the baseline Sps-\cite{vicuna68m}, which achieves a speedup ratio of 1.23. These results highlight the robustness and generalization capability of our approach in evaluation of held-out in-distribution dataset.

